\documentclass[runningheads]{llncs}
\usepackage[T1]{fontenc}
\usepackage{graphicx}
\usepackage{verbatim}
\usepackage{hyperref}
\usepackage{color}

\usepackage[table,xcdraw]{xcolor} % loads colortbl
\usepackage{amsmath}
\usepackage{amsfonts}  % mathbb
\usepackage{bm}
\usepackage{bbm}  % bold one, e.g. \mathbbm{1}
\usepackage{array}  % extends the options for column formats
\usepackage{booktabs}  % better tables: toprule, etc
\usepackage{multirow}  % \multirow command
\usepackage{makecell}  % line breaks in cells
\usepackage{caption}
\usepackage{subcaption}
\usepackage{bbding}  % \Envelope
\newcolumntype{L}[1]{>{\raggedright\let\newline\\\arraybackslash\hspace{0pt}}m{#1}}
\newcolumntype{C}[1]{>{\centering\let\newline\\\arraybackslash\hspace{0pt}}m{#1}}
\newcolumntype{R}[1]{>{\raggedleft\let\newline\\\arraybackslash\hspace{0pt}}m{#1}}

\newcommand{\realcolon}[1]{REAL-Colon}
\begin{document}
\title{Temporally-Aware Supervised Contrastive Learning for Polyp Counting in Colonoscopy}
\author{
Luca Parolari\index{Parolari, Luca}\inst{1}\Envelope\orcidID{0000-0001-8574-4997}
\and Andrea Cherubini\index{Cherubini, Andrea}\inst{2,3}\orcidID{0000-0002-5946-4390}
\and Lamberto Ballan\index{Ballan, Lamberto}\inst{1}\orcidID{0000-0003-0819-851X}
\and Carlo Biffi\index{Biffi, Carlo}\inst{2}\orcidID{0000-0002-4913-7441}
}
\authorrunning{L. Parolari et al.}
\titlerunning{Temporally-Aware Supervised Contrastive Learning for Polyp Counting}
\institute{
Department of Mathematics, University of Padova, Padova, Italy
\email{luca.parolari@phd.unipd.it}
\and Cosmo Intelligent Medical Devices, Dublin, Ireland 
\and Milan Center for Neuroscience, University of Milano–Bicocca, Milan, Italy
}

\maketitle              % typeset the header of the contribution
\begin{abstract}

Automated polyp counting in colonoscopy is a crucial step toward automated procedure reporting and quality control, aiming to enhance the cost-effectiveness of colonoscopy screening. 
Counting polyps in a procedure involves detecting and tracking polyps, and then clustering tracklets that belong to the same polyp entity.
Existing methods for polyp counting rely on self-supervised learning and primarily leverage visual appearance, neglecting temporal relationships in both tracklet feature learning and clustering stages.
In this work, we introduce a paradigm shift by proposing a supervised contrastive loss that incorporates temporally-aware soft targets. 
Our approach captures intra-polyp variability while preserving inter-polyp discriminability, leading to more robust clustering. 
Additionally, we improve tracklet clustering by integrating a temporal adjacency constraint, reducing false positive re-associations between visually similar but temporally distant tracklets.
We train and validate our method on publicly available datasets and evaluate its performance with a leave-one-out cross-validation strategy. 
Results demonstrate a 2.2x reduction in fragmentation rate compared to prior approaches. 
Our results highlight the importance of temporal awareness in polyp counting, establishing a new state-of-the-art. 
Code is available at \href{https://github.com/lparolari/temporally-aware-polyp-counting}{github.com/lparolari/temporally-aware-polyp-counting}.


\end{abstract}

\section{Introduction}

Polyp counting in colonoscopy aims to determine the number of distinct polyps observed during the entire procedure. 
It involves detecting polyps in video frames, tracking their appearances over time to form tracklets—sequences of consecutive detections, and associating together tracklets that belong to the same polyp entity~\cite{intratorSelfsupervisedPolypReidentification2023,parolari2025polypcountingfullprocedurecolonoscopy}.
Once associations are established, the polyp count can be derived, enabling the computation of quality metrics such as Polyps Per Colonoscopy (PPC) or, by adding polyp classification, the Adenoma Detection Rate (ADR).
Successfully addressing this challenge would facilitate automated report generation, enhance cost-efficiency, and support endoscopist skills assessment~\cite{gimeno2023artificial,amano2018number,lux2023assisted,tavanapong2022artificial}.

In contrast to polyp detection and tracking \cite{nogueira-rodriguezDeepNeuralNetworks2021,biffi2022novel,nie2024artificial,biffi2024real}, the task of polyp counting has received limited attention, with only two studies published to date~\cite{intratorSelfsupervisedPolypReidentification2023,parolari2025polypcountingfullprocedurecolonoscopy}. 
These studies assume tracklets as given and focus on learning a robust visual representation for clustering.
In~\cite{intratorSelfsupervisedPolypReidentification2023}, authors propose a contrastive learning approach based on the SimCLR framework~\cite{chenSimpleFrameworkContrastive2020}. 
They generate positive samples by automatically splitting tracklets into segments to create two ``views'' of the same polyp. 
In~\cite{parolari2025polypcountingfullprocedurecolonoscopy}, authors refine this approach by integrating curriculum learning
%~\cite{DBLP:conf/icml/BengioLCW09}
that gradually integrates harder samples, and introduce Affinity Propagation~\cite{frey2007clustering} clustering to enhance polyp counting.
Recent studies~\cite{DBLP:conf/icmcs/ChenCCYQX23,DBLP:conf/icassp/XiangLRCDQ24} have investigated polyp retrieval through self-supervision and contrastive learning, defining the task as matching a query polyp against a large gallery of candidates.
Unfortunately, this research has been limited to a private dataset comprising multiple colonoscopies from the same patients.

We note that all previous works leverage self-supervision for representation learning. 
While learning object semantics through positive views from data augmentation has proven highly effective for standard imaging scenario~\cite{heMomentumContrastUnsupervised2020,he2022masked,caron2021emerging,grill2020bootstrap}, its application in colonoscopy is hindered by the severe variations in both inter- and intra-polyp appearance
due to motion blur, lighting, occlusions, camera distance, debris, and intermittent visibility of polyps. In this paper, we make a two-fold contribution to overcome this issue. 

First, we learn both intra-polyp invariance and inter-polyp separability leveraging polyp entity information.
Specifically, we adopt a supervised contrastive loss~\cite{khosla2020supervised} where multiple positives samples are considered. 
% We derived positive views from polyp entities. 
This loss allows to explicitly model sample's relationships in the embedding space and promotes stronger invariance among views of the same polyp, while still ensuring sufficient separability. 
Additionally, using multiple positive samples reduces the reliance on complex hard positive or negative mining strategies~\cite{tian2023stablerep}, such as the curriculum learning approach adopted in previous work~\cite{parolari2025polypcountingfullprocedurecolonoscopy}.

Second, we integrate the temporal information intrinsic in video data~\cite{DBLP:conf/cvpr/QianMG0WBC21}.
By leveraging temporal coherence, we aim to enhance intra-polyp invariance.
Specifically, we hypothesize that enforcing similarity between temporally close tracklets, in particular for those exhibiting visual dissimilarity, will refine the learned embedding space. 
Inspired by soft-contrastive learning on natural images~\cite{sobal2024mathbb,lee2023soft}, we modify the loss function to weight targets based on temporal distance.
Moreover, we further integrate temporal awareness in the clustering stage as a penalty term. The objective is to penalize associations between visually similar tracklets that are unlikely to belong to the same polyp due to their large temporal gap.

We conduct an extensive experimental investigation on public datasets. 
While \cite{intratorSelfsupervisedPolypReidentification2023} uses a private dataset, \cite{parolari2025polypcountingfullprocedurecolonoscopy} proposes train, validation and test splits on the publicly available \realcolon{} dataset.
Due to limited availability of data we introduce two changes. First, we expand the proposed training set with three open-access datasets: LDPolyp~\cite{ma2021ldpolypvideo}, SUN~\cite{misawa2021development} and PolypSet~\cite{li2021colonoscopy}, totalying 384 polyps (299 more than in previously proposed split). For fair comparison, we re-train previous methods.
Then, we transition from a simple validation and test split to a thoughtful evaluation using leave-one-out cross-validation (LOOCV).
Results demonstrate that our method is the state-of-the-art in polyp counting obtaining an improvement of 2.2x with respect to previous works.

In summary, this work focuses on tracklet representation learning and clustering for polyp counting. Our main contributions include: (1) a shift from self-supervised to supervised contrastive learning, improving inter-polyp separability; (2) a contrastive loss with temporally-aware soft targets for better intra-polyp invariance; (3) a tracklet clustering method using temporal penalties to discourage unlikely associations; and (4) a comprehensive benchmark against competing methods, leveraging an expanded dataset and a robust LOOCV evaluation.

\section{Method}

\begin{figure}[ht]
\centering
\begin{subfigure}[t]{1\textwidth}
  \centering
  \includegraphics[width=0.9\textwidth]{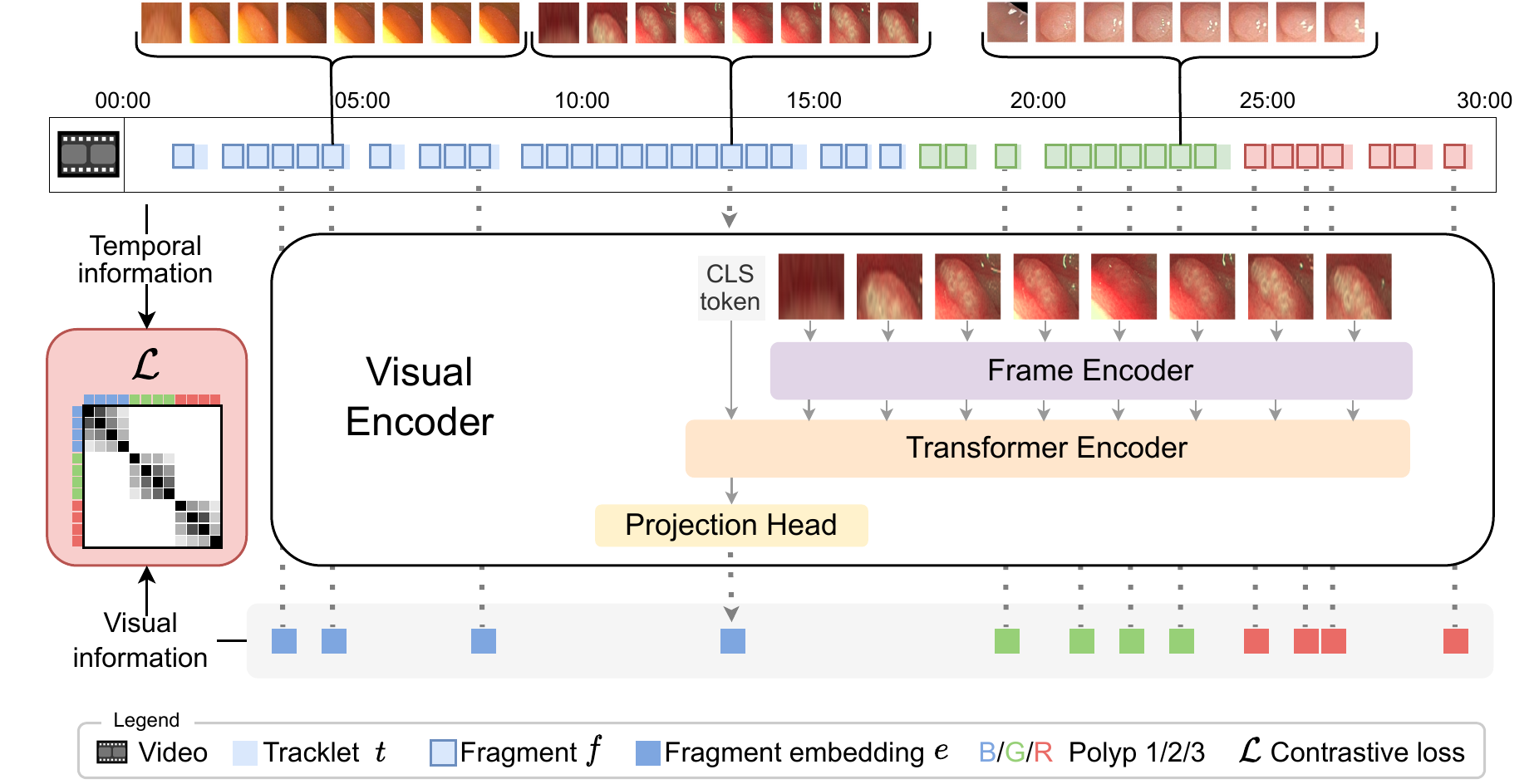}
  \label{fig:method-encoder}
\end{subfigure}
\par\bigskip % force a bit of vertical whitespace
\begin{subfigure}[t]{1\textwidth}
  \centering
  \includegraphics[width=0.9\textwidth]{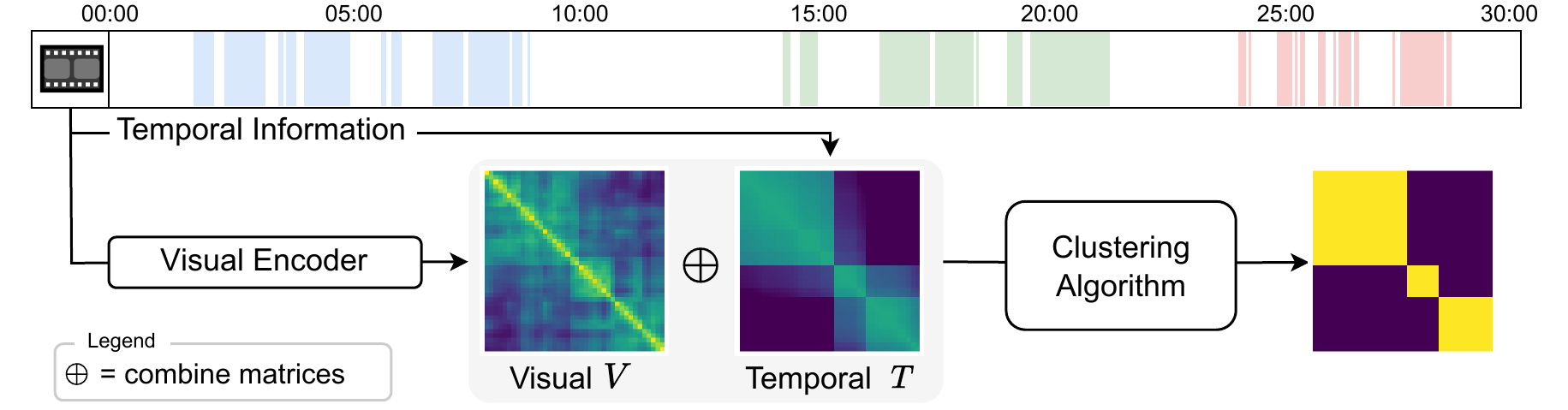}
  \label{fig:method-clustering}
\end{subfigure}
\caption{Our method. \textit{(Top)} Polyp tracklets detected in a colonoscopy video are processed by the encoder to extract their visual-temporal feature representation. The encoder is trained with a supervised contrastive loss using temporally-aware soft targets, computed on batches that include multiple views of multiple polyps, potentially from different videos. \textit{(Bottom)} The clustering module obtains the tracklet embeddings and computes the similarity matrix. This matrix is combined with the temporal adjacency matrix, and a clustering algorithm derives the set of polyp entities.}
\label{fig:method}
\end{figure}

Our approach is based on two main steps.
First, the visual-temporal encoder, depicted in Fig.~\ref{fig:method}~(Top), obtains tracklet representation.
We train the encoder with a novel temporally-aware supervised contrastive loss, where soft-targets are introduced to reflect temporal proximity between tracklets.
This encourages embeddings of temporally adjacent but visually dissimilar samples to be considered similar while maintaining class discriminability.
Then, we employ a clustering module to re-associate tracklets into polyp entities and count them. We represent this module in Fig.~\ref{fig:method}~(Bottom).
This step leverages both the visual features extracted by the encoder and temporal information derived from the video.
Specifically, we integrate a temporal penalty term to penalize the association of distant tracklets, thereby reducing false positive re-associations that occur between temporally distant tracklets in the video.

\subsection{Learning temporally-aware polyp representation}

The visual-temporal encoder is responsible for generating tracklets representation, and is depicted in Fig.~\ref{fig:method} (Top). Its architecture is based on~\cite{intratorSelfsupervisedPolypReidentification2023} and involves two components: a frame-level feature extractor and sequence-level aggregator.
The frame-level feature extraction is implemented by a visual backbone, e.g. ResNet~\cite{he2016deep}, to capture the appearance of polyps in individual frames. 
The frame embeddings are then processed by a transformer network~\cite{vaswani2017attention} to capture frame dependencies. The final tracklet representation is obtained from the classification token (CLS) embedding~\cite{devlinBERTPretrainingDeep2019}, and projected by a non-linear head~\cite{chenSimpleFrameworkContrastive2020}.
Specifically, the encoder receive a fragment $\bm{f}_{i} = \bm{t}_{[i\cdot\kappa:i\cdot\kappa+\kappa]}, i \in [0, \ldots, M]$ and returns its embedding $\bm{e}_i \in \Re^d$. Here, $\bm{t}$ is a tracklet from video $\mathcal{V}$, and $M = \lfloor N/\kappa \rfloor$, with $N$ the actual length of the tracklet.
Thus, each fragment encodes $\kappa$ images $\bm{f}_{i} = [\bm{x}^1_i, \ldots, \bm{x}^\kappa_i]$. 
Each image $\bm{x}^j_i$ represents a crop to the bounding box of the polyp instance.
To include context around the polyp, we enlarge the bounding box to $\psi$ times its diagonal before cropping.

\begin{figure}[t!]
\centering
\includegraphics[width=0.9\textwidth]{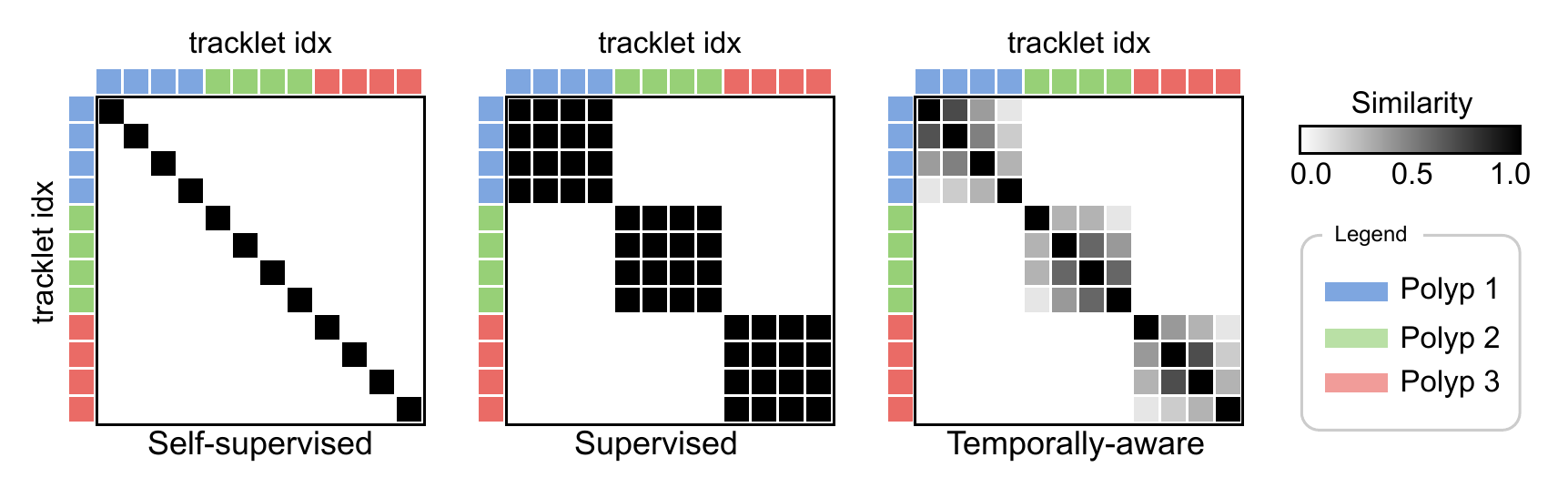}
\caption{Comparison of similarity adjacency matrices: self-supervised (left), supervised (center), and our temporally-aware contrastive loss (right). We visualize pairwise similarities for 12 tracklets from 3 polyps. Self-supervised learning does not explicitly define relationships between samples, supervised learning exploits polyp entity information to improve inter-polyp features modeling, while our approach improves also intra-polyp features by integrating temporal information.}
\label{fig:loss}
\end{figure}

We train the encoder with the contrastive loss defined in~\cite{khosla2020supervised}, which extends the normalized temperature-scaled cross-entropy loss used in SimCLR~\cite{chenSimpleFrameworkContrastive2020}. 
This loss considers multiple positives per anchor, rather than just a single one, allowing the model to explicitly define inter-sample relationships in the embedding space. 
Instead of equally representing all positive views in the ground truth distribution, we weight each view by its temporal distance with respect to the anchor.
We depict our approach compared to traditional self-supervised and supervised in Fig.~\ref{fig:loss}.
This approach promotes intra-polyp consistency without sacrificing the ability to distinguish between different polyp instances and reduces sensitivity to appearance variation.

Formally, we formulate the loss as a matching problem~\cite{tian2023stablerep}, where an encoded anchor sample $\bm{a}$ is compared against a set of candidates $\{\bm{b}_1, \bm{b}_2, ..., \bm{b}_K\}$.
The probability of $\bm{a}$ matching each candidate $\bm{b}_i$ is modeled using a softmax distribution:
\begin{equation}
\bm{q}_i = \frac{\exp(\bm{a} \cdot \bm{b}_i / \tau)}
{\sum_{j=1}^{K} \exp(\bm{a} \cdot \bm{b}_j / \tau)}
\end{equation}
where $\tau$ is a temperature parameter, and all vectors have been $\ell_2$ normalized. 
To introduce temporal awareness in the ground-truth distribution, we obtain a vector $\bm{d}$ that encodes the temporal distances between $\bm{a}$ and all $\bm{b}_i$, i.e.
$\bm{d} = |\bm{\pi} - \pi| / P$ where $\pi$ and $\bm{\pi}$ represent the timestamps of $\bm{a}$ and all $\bm{b}_i$ in the video, $P$ is the length of the polyp to which the anchor belongs to. 
The ground-truth distribution is defined as:
\begin{equation}
\bm{p}_i = \frac{\mathbbm{1}_{\text{match}}(\bm{a}, \bm{b}_i) \cdot \exp(-\lambda \bm{d}_i)}
{\sum_{j=1}^{K} \mathbbm{1}_{\text{match}}(\bm{a}, \bm{b}_j) \cdot \exp(-\lambda \bm{d}_j)}
\end{equation}
where $\mathbbm{1}_{\text{match}}(\cdot, \cdot)$ is an indicator function identifying matches and $\lambda$ is a scaling factor that controls the influence of temporal distance. 
This formulation ensures that matching candidates closer in time to $\bm{a}$ receive higher weights independently of their visual appearance, while more distant matches are exponentially suppressed. We provide a visualization in Fig.~\ref{fig:loss}~(right).
The loss is then given by the cross-entropy between $\mathbf{p}$ and $\mathbf{q}$:
\begin{equation}
\mathcal{L} = H(\mathbf{p}, \mathbf{q}) = - \sum_{i=1}^{K} p_i \log q_i.
\end{equation}

\subsection{Polyp counting via clustering with temporal penalties}

Similarly to~\cite{parolari2025polypcountingfullprocedurecolonoscopy}, this module aims to cluster together tracklets that correspond to the same polyp entity.
Alongside to visual similarity, we introduce another measure of similarity based on temporal information. Specifically, we integrate a temporal penalty term that penalizes associations between tracklets that are far apart in time, and thus are unlikely to belong to the same cluster. The Fig.~\ref{fig:method}~(Bottom) depicts this component in details.

Formally, given a set of tracklet $\{\bm{t}_i\}$ from a video with $N$ tracklets, we compute the embedding $\bm{e}_i$ for each of them through the visual-temporal encoder. Then, we obtain $\bm{p}_i$, the vector that encodes the position of $\bm{t}_i$ against all $\bm{t}_j$ in video, normalized by video length.
In order to perform the clustering, we obtain two matrices. The first is a visual similarity score matrix $V \in \mathbb{R}^{N \times N}$, that models appearance-based similarity. $V$ is obtained by computing all pairwise similarity scores $V_{i,j} = \text{sim}(\bm{e}_i, \bm{e}_j)$, where $\text{sim}$ is a similarity function, e.g. cosine similarity.
The second is a temporal adjacency matrix $T \in \mathbb{R}^{N \times N}$, $T_{i,j} = e^{-\gamma |\bm{p}_i - \bm{p}_j|}$ where $\gamma$ is the penalty factor. We normalize both $V$ and $T$ to get values in $[0, 1]$.

Finally, we compute $S = V \oplus T$ by combining $V$ and $T$, e.g. by summing the contributions: $V \oplus T = \alpha \cdot V + (1-\alpha) \cdot T$ and clustering can be performed. 
Literature associates tracklets together either by using a threshold based approach~\cite{intratorSelfsupervisedPolypReidentification2023} or unsupervised clustering algorithms~\cite{parolari2025polypcountingfullprocedurecolonoscopy}.
Given the superior performance shown by the latter method, especially by Affinity Propagation~\cite{frey2007clustering}, we adopt this algorithm for our approach.
From clusters, polyp counting can be finally derived.

\section{Experiments}

\subsection{Datasets and evaluation protocol}

% \paragraph{Datasets and evaluation protocol.} 
We construct our training set from publicly available datasets, and follow the evaluation framework proposed in~\cite{parolari2025polypcountingfullprocedurecolonoscopy}. 
Specifically, we extend the training set by combining 85 polyps from \realcolon{}, already used in~\cite{parolari2025polypcountingfullprocedurecolonoscopy}, with 160 from LDPolyp~\cite{ma2021ldpolypvideo}, 100 from SUN~\cite{misawa2021development} and 39 from PolypSet~\cite{li2021colonoscopy}, for a total of 384 polyps. 
Each dataset provides polyp detections, but (polyp, video) annotations are available only in \realcolon{}. 
For this reason, we cannot extend the evaluation set to other datasets. In~\cite{parolari2025polypcountingfullprocedurecolonoscopy}, authors propose a validation set with 10 videos (23 polyps) and a test set with 9 videos (24 polyps).
Although REAL-Colon is a multi-centric dataset, the limited size of the validation set may lead to unreliable estimates of test performance.
To address this, we use a leave-one-out cross-validation (LOOCV) strategy, ensuring a more reliable and robust assessment.
Following~\cite{intratorSelfsupervisedPolypReidentification2023,parolari2025polypcountingfullprocedurecolonoscopy}, we measure the fragmentation rate (FR), defined as the average number of tracklets polyps are split into. 
The FR is a number $\geq 1$, where lower values mean lower fragmentation. 
Formally, we define the FR as a function of $T$ a set of tracklets and $E$ a set of polyp entities. 
Given a set of clustered tracklets $C$, with $|C| \leq |T|$, we can compute the fragmentation rate as $FR = |C| / |E|$. A perfect re-association would have $|C| = |E|$, thus $FR = 1$.
To account for false positive associations, we select the clustering hyper-parameters that yield the lowest fragmentation rate (FR) at fixed false positive rate (FPR) $\rho = 0.05$ on the combined validation set of 18 videos and evaluate on the remaining one, reporting the average FR and FPR across all 19 videos. 
Thus, $FR = |\hat{C}| / |E|$, where $\hat{C}$ is the set of clustered tracklets that minimizes $|FPR(C) - \rho|$.

\subsection{Implementation details}
The visual encoder is implemented following~\cite{intratorSelfsupervisedPolypReidentification2023}. For comparability with previous works~\cite{intratorSelfsupervisedPolypReidentification2023,parolari2025polypcountingfullprocedurecolonoscopy}, we use ResNet-50 as frame encoder pre-trained on ImageNet~\cite{DBLP:conf/cvpr/DengDSLL009}, and set the model size $d = 128$. Due to memory limitations, we set fragment length $\kappa = 8$ and batch size to $56$. Number of different polyps in a batch is set to 3, i.e. 14 views per polyp (unless otherwise specified). We set $\lambda = 1$. 
We select the bounding box scaling factor $\psi$ in $\{1, 2, 3, 5, 10\}$ and chose $5$. For fair comparison, all methods were trained and evaluated under the same setting.
We construct tracklets from ground-truth polyp detections to avoid tracker noise and enable fair comparison with prior work. Each tracklet is a sequence of consecutive frames from the same polyp entity, with an Intersection Over Union (IoU) of at least $0.1$ between consecutive frames. 
To avoid redundancy from high fps videos in tracklets, we encode one frame every four.
We train all models on the combined dataset from public sources. Our runtime machine is a node with 6 cores, 60GB of RAM and one NVIDIA RTX A6000 GPU with 48GB of VRAM.
We select clustering hyper-parameters with a grid-search. For the threshold algorithm we vary the threshold in $[0, 0.01, \ldots, 1]$. For Affinity Propagation we select the preference parameter in $[-5, -4.75, \ldots, 5]$, while for our temporal clustering we select $\gamma \in [0.1, 0.2, \ldots, 0.9] \cup [1, 1.375, \ldots, 10]$ and $\alpha \in [0, 0.05, \ldots, 1]$ along with Affinity Propagation's preference.

\subsection{Results}

\begin{table}[t]
\centering
\begin{tabular}{l|cc}
\toprule
Method & FR $\downarrow$ & FPR $\downarrow$ \\
\midrule
No clustering & 20.74 $\scriptstyle{\pm 20.78}$ & 0.0000 $\scriptstyle{\pm 0.0000}$ \\ 
\hline
SFE~\cite{intratorSelfsupervisedPolypReidentification2023} & 9.37 $\scriptstyle{\pm 6.28}$ & 0.0314 $\scriptstyle{\pm 0.0618}$ \\
MVE~\cite{intratorSelfsupervisedPolypReidentification2023}  & 8.42 $\scriptstyle{\pm 4.50}$  & 0.0449 $\scriptstyle{\pm 0.1361}$ \\
TPC~\cite{parolari2025polypcountingfullprocedurecolonoscopy} & 5.10 $\scriptstyle{\pm 3.43}$  & 0.0249 $\scriptstyle{\pm 0.0408}$ \\
\textbf{Ours} & \textbf{2.32} $\bm{\scriptstyle{\pm 1.80}}$  & \textbf{0.0225} $\bm{\scriptstyle{\pm 0.0420}}$ \\
\bottomrule
\end{tabular}
\caption{Mean and standard deviation results of the Fragmentation Rate (FR) and False Positive Rate (FPR) on REAL-Colon using LOOCV. We report results for models trained on the combination of public datasets composed by \realcolon{}, LDPolyp, SUN and PolypSet. The ``No clustering'' baseline reflects the initial fragmentation rate before any clustering.}
\label{tab:results}
\end{table}

We compare our approach to the state-of-the-art~\cite{intratorSelfsupervisedPolypReidentification2023,parolari2025polypcountingfullprocedurecolonoscopy} and present results in Tab.~\ref{tab:results}.
For fair comparison, we train previous methods on the same combined dataset used in our experiments. 
Our method outperforms all existing approaches, achieving 3.6 and 4.0 times lower fragmentation rate with respect to SFE and MVE from~\cite{intratorSelfsupervisedPolypReidentification2023}.
Notably, TPC~\cite{parolari2025polypcountingfullprocedurecolonoscopy} also employs Affinity Propagation for clustering, the same underlying algorithm used in our approach, but our method yields a 2.2x improvement thanks to the combination of our novel visual-temporal representation learning and temporal clustering.

\subsection{Ablation study}

\paragraph{Effect of multiple positive views per polyp on representation learning.}
Previous works~\cite{intratorSelfsupervisedPolypReidentification2023,parolari2025polypcountingfullprocedurecolonoscopy} follows the traditional contrastive learning framework~\cite{chenSimpleFrameworkContrastive2020}, adopting two positive views per polyp. 
Instead, our supervised setting allows to derive multiple meaningful views from polyp entities. 
We evaluate the impact of this parameter on the quality of representations. 
With a fixed batch capacity of 56 elements, we investigate the number of views per polyp $K$ in $\{2, 4, 7, 14, 28\}$, corresponding to $56/K = 28$, 12, 8, 3, and 2 polyps per batch, respectively. 
Our results show that increasing the number of views helps reducing the fragmentation rate (FR) and false positive rate (FPR), with the lowest FR (2.32) and FPR (0.0225) achieved with 14 views (3 polyps in batch).

\begin{table}[t]
\centering
\begin{tabular}{cccc}
\toprule
N. views per polyp & N. polyps in batch & FR $\downarrow$ & FPR $\downarrow$ \\
\midrule
2 & 28 & 2.57 $\scriptstyle{\pm 2.07}$ & 0.0276 $\scriptstyle{\pm 0.0576}$ \\
4 & 14 & 3.97 $\scriptstyle{\pm 3.13}$ & 0.0417 $\scriptstyle{\pm 0.0987}$ \\
7 & 8 & 2.36 $\scriptstyle{\pm 1.77}$ & 0.0325 $\scriptstyle{\pm 0.0642}$ \\
\textbf{14} & \textbf{3} & \textbf{2.32} $\bm{\scriptstyle{\pm 1.80}}$ & \textbf{0.0225} $\bm{\scriptstyle{\pm 0.0420}}$ \\
28 & 2 & 4.54 $\scriptstyle{\pm 3.86}$ & 0.0317 $\scriptstyle{\pm 0.0557}$ \\
\bottomrule
\end{tabular}
\caption{Fragmentation rate results on REAL-Colon by number of views per polyp in a batch of 56 elements.}
\label{tab:ablation-views}
\end{table}

\paragraph{Impact of temporal information on polyp counting.}

We investigate the impact of temporal information on both tracklet representation learning and clustering in Tab.~\ref{tab:ablation-loss}. On the top, we can note the superior performance of our temporally-aware supervised loss over the standard supervised loss, which in turn improves on the self-supervised loss.
Since the clustering algorithm remains unchanged, the observed improvement can be attributed to a more discriminative embedding space.
On the bottom, the results show that incorporating temporal information into the clustering process (Temporal-AP) significantly reduces fragmentation rate (FR) suggesting that the temporal association penalty helps form more accurate clusters. 
Notably, both visual and temporal cues contribute to the final clustering. The weighting factor $\alpha$, which balances the influence of the two modalities, averages $0.4833 \pm 0.1999$ across the LOOCV folds, indicating that their combination is essential for optimal performance.

\begin{table}[t]
\centering
\begin{tabular}{ll|cc}
\toprule
Loss & Clustering & FR $\downarrow$ & FPR $\downarrow$ \\
\midrule
Self-supervised & Temporal-AP & 10.47	$\scriptstyle{\pm 19.96}$ & 0.0210 $\scriptstyle{\pm 0.0248}$ \\
Supervised & Temporal-AP & 3.96 $\scriptstyle{\pm 2.68}$ & 0.0252 $\scriptstyle{\pm 0.0507}$ \\ 
\textbf{Temporally-aware} & \textbf{Temporal-AP} & \textbf{2.32} $\bm{\scriptstyle{\pm 1.80}}$  & \textbf{0.0225} $\bm{\scriptstyle{\pm 0.0420}}$ \\
\midrule
Temporally-aware & AP & 4.18 $\scriptstyle{\pm 2.86}$ & 0.0259 $\scriptstyle{\pm 0.0397}$ \\
\textbf{Temporally-aware} & \textbf{Temporal-AP} & \textbf{2.32} $\bm{\scriptstyle{\pm 1.80}}$ & \textbf{0.0225} $\bm{\scriptstyle{\pm 0.0420}}$ \\
\bottomrule
\end{tabular}
\caption{Ablation results on \realcolon{} by loss functions and clustering algorithms. AP=Affinity Propagation, Temporal-AP=AP with temporal penalty.}
\label{tab:ablation-loss}
\end{table}

\section{Conclusion}

In conclusion, we propose a novel polyp counting approach using supervised contrastive learning to enhance tracklet representation. By incorporating temporally-aware soft targets, our method improves intra-polyp invariance and inter-polyp separability. Temporal constraints in the clustering stage further refine tracklet associations, enhancing reliability. Extensive evaluation on public datasets, including an expanded training set and leave-one-out cross-validation, shows that our method outperforms previous approaches, reducing fragmentation by a factor of 2.2.

\begin{credits}
\subsubsection{\ackname}
We acknowledge ISCRA for awarding this project access to the LEONARDO supercomputer, owned by the EuroHPC Joint Undertaking, hosted by CINECA (Italy).

\subsubsection{\discintname}
C.B. and A.C. are affiliated with Cosmo Intelligent Medical Devices, the developer of the GI Genius medical device.
\end{credits}

\bibliographystyle{splncs04}
\bibliography{main}

\end{document}